\documentclass[runningheads]{llncs}
\usepackage[T1]{fontenc}
\usepackage{graphicx}
\usepackage{booktabs}
\usepackage[misc]{ifsym}

\usepackage{fvextra}

\begin{document}

\title{On The Dynamic Ensemble Selection for TinyML-based Systems - a Preliminary Study}
\titlerunning{On The Dynamic Ensemble Selection for TinyML}
\author{Tobiasz Pu\'slecki\thanks{Corresponding Author. Email: tobiasz.puslecki@pwr.edu.pl}\inst{1} \and Krzysztof Walkowiak\inst{1}}
\authorrunning{T. Puślecki and K. Walkowiak}
\institute{$^{1}$Department of Systems and Computer Networks, \\Wrocław University of Science and Technology}

\maketitle

\begin{abstract}

The recent progress in TinyML technologies triggers the need to address the challenge of balancing inference time and classification quality. TinyML systems are defined by specific constraints in computation, memory and energy. These constraints emphasize the need for specialized optimization techniques when implementing Machine Learning (ML) applications on such platforms. While deep neural networks are widely used in TinyML, the exploration of Dynamic Ensemble Selection (DES) methods is also beneficial. This study examines a \verb|DES-Clustering| approach for a multi-class computer vision task within TinyML systems. This method allows for adjusting classification accuracy, thereby affecting latency and energy consumption per inference. We implemented the \verb|TinyDES-Clustering| library, optimized for embedded system limitations. Experiments have shown that a larger pool of classifiers for dynamic selection improves classification accuracy, and thus leads to an increase in average inference time on the TinyML device.

\end{abstract}

\section{Introduction}

TinyML involves implementing machine learning models on microcontrollers or other devices with limited resources. In many TinyML applications, power consumption is a critical constraint. Collecting and processing data in an energy-efficient manner while maintaining model accuracy is a major challenge. TinyML models must be lightweight and efficient to run on resource-constrained devices. Balancing model complexity with performance is critical, often requiring the use of specialized algorithms and optimization techniques tailored for implementation on microcontrollers~\cite{tinySurvey2}~\cite{tinySurvex}.

The need to explore new energy-saving methods forces the exploration of methods other than Deep Neural Networks (DNN). An alternative to deep learning methods could be ensemble methods~\cite{ensemble}. In ensemble methods, many weak models are trained offline (assumption: resources for training unlimited), and at execution time the system selects the appropriate models based on certain conditions. For example, fewer models may be used when resources are limited, while a more complex ensemble may be selected when sufficient memory and computing power are available. In selecting classifiers for an ensemble, diversity is the key. Diversity can be achieved by selecting classifiers from different model families (e.g., decision trees, linear models) or different sets of hyperparameters. Diversity can help reduce the risk of overfitting and improve the ensemble ability to generalize. 

Moreover, it seems reasonable to assume that the more classifiers there are in an ensemble, the longer it takes to process an ensemble, leading to the consumption of more energy. Since the underlying classifiers are relatively weak, it is reasonable to conclude that more classifiers in an ensemble leads to higher efficiency. Thus, by varying the number of classifiers in an ensemble, a tradeoff can be made between accuracy and energy consumption. 
For example, for a hard-to-reach battery-powered system, it will be possible to increase the operating time without significant loss of accuracy. 
In this paper, we demonstrate a novel application of the \verb|DES-Clustering|\footnote{\url{https://github.com/tobiaszpuslecki/TinyDES}} method to the TinyML system. To the best of our knowledge, no one has combined the two fields before. We present a work in progress with the first results as a proof of concept.

\section{Dynamic Ensemble Selection}
Different baseline models can be competent in different local regions of the feature space for a given sample. For that reason, a convenient approach is to select the most competent models from the pool to the predicting ensemble. This approach is widely known as Dynamic Ensemble Selection (DES). In opposite to static selection, where the selection of classifiers is performed in the training phase, dynamic selection is performed for each new test sample in the classification phase. DES systems could be an alternative for increasing overall system accuracy over monolithic classifiers like DNNs. Classifiers ensembles are widely used to solve many real-world problems, such as face recognition, music genre classification, intrusion detection, and for dealing with changing environments~\cite{2a}.

The most popular DES methods are neighborhood-based (e.g. \verb|KNORA-U/E|) and clustering-based (e.g. \verb|DES-Clustering|). Neighborhood-based dynamic selection methods are not a good choice for resource-constrained systems -- they do not require training, but it is necessary to store in ROM a validation set for assessing regions of competence. Although, the naive version of the \textit{k}NN algorithm is easy to implement by computing the distances from the test example to all stored examples, but it is computationally intensive for large training sets. That's the reason why dynamic selection methods based on clustering are interesting in terms of embedded systems. Then only \textit{N} centroids are stored in ROM with pools of base classifiers assigned to them. \verb|DES-Clustering| method selects an ensemble of classifiers taking into account the accuracy and diversity of the base classifiers. The \textit{K}-means algorithm is used to define the region of competence. For each cluster, the \textit{N} most accurate classifiers are first selected. Then, the \textit{J}~more diverse classifiers from the \textit{N} most accurate classifiers are selected to compose the ensemble. Since the value of \textit{J} expresses the number of classifiers used per inference, it can be hypothesized that the value of \textit{J} is directly proportional to the time of inference and thus the energy consumed. In addition, a higher number of classifiers per inference gives higher confidence in prediction, which leads to the hypothesis that the value of \textit{J} allows to regulate the relationship between prediction efficiency and energy consumption. Clustering-based dynamic selection methods, in contrast to neighborhood-based methods, are heavier at the training stage, but lighter at the inference stage. Due to limited resources, it should be assumed that both the clustering method and the base classifiers should have memory and computational complexity (the pool of classifiers must fit in memory) befitting TinyML systems. 

\section{Experiments}

All experiments are implemented in Python, using scikit-learn~\cite{scikit-learn}, numpy~\cite{numpy} and DESLib~\cite{deslib} libraries. Three computer vision datasets are used, MNIST~\cite{mnist}, Fashion-MNIST~\cite{fashion} and EMNIST~\cite{emnist}, which are flattened and standardized. The models are cross-validated times times using the RepeatedStratifiedKFold method. The results of each cross-validation are averaged. A very important aspect in dynamic selection is the generation of a pool of classifiers. A common practice in the dynamic selection literature is to use the Bagging (Bootstrap Aggregating) method to generate a pool containing base classifiers that are both diverse and informative. In this research we use a pool of classifiers generated using the Random Forest method. Random Forest is a well-known algorithm, which combines the concepts of bagging and random subspace. To ensure diversity in the ensemble using homogeneous classifiers, two Random Forests differing in hyperparameters (here depth) are used. The pool of base classifiers consists of an RF containing 25 estimators with a maximum depth of 10 and an RF containing 20 estimators with a maximum depth of 5. \textit{K}-means algorithm is used to define the region of competence (with \textit{k}=5). In this experiment, we vary the value of parameter \textit{J}~and measure the performance of 6 different selection techniques using the same pool of classifiers.

We compare the performance of dynamic selection techniques with the \Verb|Single| \Verb|Best|, \Verb|Oracle| and \Verb|Static Selection| methods. The following techniques are considered:

\begin{itemize}
    \item \textbf{Single Best} - The base classifier with the highest classification accuracy in the validation set is selected for classification~\cite{1a,2a,4a}.
    \item \textbf{Static Selection} - A fraction of the best performing classifiers, based on the validation data, is selected to compose the ensemble -- ensemble couldn't be changed in runtime~\cite{1a,2a,4a}.
    \item \textbf{k-Nearest Oracles Union} (\verb|KNORA-U|) - This method selects all classifiers that correctly classified at least one sample belonging to the region of competence of the query sample. Each selected classifier has a number of votes equals to the number of samples in the region of competence that it predicts the correct label. The votes obtained by all base classifiers are aggregated to obtain the final ensemble decision~\cite{1a,2a,3a}.
    \item \textbf{k-Nearest Oracles Eliminate} ($\verb|KNORA-E|$) - This method searches for a~local Oracle, which is a base classifier that correctly classify all samples belonging to the region of competence of the test sample. All classifiers with a perfect performance in the region of competence are selected (local Oracles). In the case that no classifier achieves a perfect accuracy, the size of the competence region is reduced (by removing the farthest neighbor) and the performance of the classifiers are re-evaluated. The outputs of the selected ensemble of classifiers is combined using the majority voting scheme. If no base classifier is selected, the whole pool is used for classification~\cite{1a,2a,3a}.
    \item \textbf{Dynamic Ensemble Selection-Clustering} (\verb|DES-Clustering|) - This me-thod selects an ensemble of classifiers taking into account the accuracy and diversity of the base classifiers. The \textit{K}-means algorithm is used to define the region of competence. For each cluster, the \textit{N} most accurate classifiers are first selected. Then, the \textit{J} more diverse classifiers from the \textit{N} most accurate classifiers are selected to compose the ensemble~\cite{1a,5a,6a}.
    \item \textbf{Oracle} - Abstract method that always selects the base classifier that predicts the correct label if such classifier exists. This method is often used to measure the upper-limit performance that can be achieved by a dynamic classifier selection technique. It is used as a benchmark~\cite{5a,7a}.
\end{itemize}

\begin{table}[ht!]
\centering
\caption{The values of accuracy metric depending on dataset and selected method.}\label{tab1}
\begin{tabular}{l|ccc|}
\cline{2-4}
& \multicolumn{3}{c|}{Overall Accuracy (std)}                                                           \\ \cline{2-4} 
& \multicolumn{1}{c|}{MNIST}         & \multicolumn{1}{c|}{Fashion MNIST} & \multicolumn{1}{c|}{EMNIST} \\ \hline
\multicolumn{1}{|l|}{\Verb|Single Best|}                             & \multicolumn{1}{c|}{0.763 (0.004)} & \multicolumn{1}{c|}{0.762 (0.005)} & 0.491 (0.008)               \\ \hline
\multicolumn{1}{|l|}{\Verb|Static Selection|}                        & \multicolumn{1}{c|}{0.927 (0.003)} & \multicolumn{1}{c|}{0.842 (0.001)} & 0.727 (0.002)               \\ \hline
\multicolumn{1}{|l|}{\Verb|KNORA-U| (\textit{k}=7)} & \multicolumn{1}{c|}{0.936 (0.002)} & \multicolumn{1}{c|}{0.843 (0.002)} & 0.754 (0.001)               \\ \hline
\multicolumn{1}{|l|}{\Verb|KNORA-E| (\textit{k}=7)} & \multicolumn{1}{c|}{0.916 (0.002)} & \multicolumn{1}{c|}{0.829 (0.002)} & 0.673 (0.003)               \\ \hline
\multicolumn{1}{|l|}{\Verb|DES-Clustering_5|}                       & \multicolumn{1}{c|}{0.879 (0.002)} & \multicolumn{1}{c|}{0.822 (0.002)} & 0.633 (0.002)               \\ \hline
\multicolumn{1}{|l|}{\Verb|DES-Clustering_10|}                      & \multicolumn{1}{c|}{0.910 (0.002)} & \multicolumn{1}{c|}{0.836 (0.002)} & 0.695 (0.002)               \\ \hline
\multicolumn{1}{|l|}{\Verb|DES-Clustering_15|}                      & \multicolumn{1}{c|}{0.922 (0.002)} & \multicolumn{1}{c|}{0.840 (0.002)} & 0.715 (0.002)               \\ \hline
\multicolumn{1}{|l|}{\Verb|DES-Clustering_20|}                      & \multicolumn{1}{c|}{0.926 (0.002)} & \multicolumn{1}{c|}{0.842 (0.002)} & 0.727 (0.003)               \\ \hline
\multicolumn{1}{|l|}{\Verb|Oracle|}                                  & \multicolumn{1}{c|}{0.999 (0.000)} & \multicolumn{1}{c|}{0.993 (0.001)} & 0.978 (0.002)               \\ \hline
\end{tabular}
\end{table}

\begin{table}[ht!]
\caption{The values of average inference time depending on dataset and selected method.}\label{tab2}
\centering
\begin{tabular}{l|ccc|}
\cline{2-4}
& \multicolumn{3}{c|}{Avg. Inference Time (ms)}                                                 \\ \cline{2-4} 
& \multicolumn{1}{c|}{MNIST} & \multicolumn{1}{c|}{Fashion MNIST} & \multicolumn{1}{c|}{EMNIST} \\ \hline
\multicolumn{1}{|l|}{\Verb|DES-Clustering_5|}  & \multicolumn{1}{c|}{1.853} & \multicolumn{1}{c|}{1.853}         & 1.878                       \\ \hline
\multicolumn{1}{|l|}{\Verb|DES-Clustering_10|} & \multicolumn{1}{c|}{1.898} & \multicolumn{1}{c|}{1.900}         & 1.943                       \\ \hline
\multicolumn{1}{|l|}{\Verb|DES-Clustering_15|} & \multicolumn{1}{c|}{1.960} & \multicolumn{1}{c|}{1.965}         & 2.028                       \\ \hline
\multicolumn{1}{|l|}{\Verb|DES-Clustering_20|} & \multicolumn{1}{c|}{2.040} & \multicolumn{1}{c|}{2.048}         & 2.130                       \\ \hline
\end{tabular}
\end{table}

Table~\ref{tab1} presents the values of accuracy metric depending on dataset and selected method. Statistical analysis (with the significance level $\alpha=0.05$) focusing on \verb|DES-Clustering| methods shows that \verb|DES-Clustering| is always statistically significantly better than \verb|Single| \verb|Best|. In addition, \Verb|Static Selection| is always statistically significantly better than \verb|DES-Clustering_5|, \verb|DES-Clustering_10| and \verb|DES-Clustering_15|. However, the static method pool consists of 22 base classifiers and cannot be changed at runtime. It is noteworthy that the larger the value of J, the statistically better the \verb|DES-Clustering| method performs (i.e. \verb|DES-Clustering_20| is better than \verb|DES-Clustering_15|, \verb|DES-Clustering_15| is better than \verb|DES-Clustering_10|. Furthermore, \verb|DES-Clustering_10| is better than \verb|DES-Clustering_5|). These findings confirm that the larger the pool of classifiers available for dynamic selection, the more accurate a set of classifiers can be.

In addition, a new custom \verb|TinyDES-Clustering| library is implemented, tailored to the limitations of embedded systems. Using the DESlib framework, the DESClustering method is trained. Then, using an implemented converter, this classifier is ported to C language, as it is commonly used in embedded systems. To profile the methods, the \verb|NUCLEO-L476RG| module, based on the 32-bit \verb|STM32L476RG| microcontroller, is used~\cite{stmanual}. The microcontroller is equipped with an \emph{ARM Cortex M4 core} clocked at 80 MHz, 1 MB flash memory, 128 kB SRAM. Table~\ref{tab2} presents the values of average inference time (including selector and classifier pool operations) depending on dataset and selected method.

Measurements confirm that the larger the pool of classifiers available for dynamic selection, the higher the average inference time. In addition, although the three datasets have equal resolution, there is an apparent difference in the values of the average inference time.

\section{Conclusion}

In this paper, we have investigated the \verb|DES-Clustering| method for a multi-class computer vision task in TinyML systems. This method allows us to adjust the classification accuracy, thereby affecting latency and energy consumption for inference. We implemented the \verb|TinyDES-Clustering| library, optimized for the limitations of embedded systems. Tests on three data sets showed that a larger pool of classifiers for dynamic selection improves the accuracy of the classifier sets. However, our measurements also indicate that increasing the pool size leads to higher average inference times of a TinyML device. \verb|TinyDES-Clustering| library can potentially be used in any TinyML application where balancing accuracy and energy consumption is crucial. In future work, we plan to extend the presented concept by incorporating battery level dependency and integrating energy harvesting via solar panels. This will enable the system to recharge the battery and dynamically adjust the value of the \textit{J} in real time. Further work may also include exploring additional use cases.

\bibliographystyle{splncs04}

\end{document}